\documentclass[conference]{IEEEtran}
\IEEEoverridecommandlockouts
\usepackage{graphicx} 
\usepackage{amsmath}
\usepackage{xcolor}
\usepackage{multirow}
\usepackage{tabularx}
\usepackage{adjustbox}

\usepackage{algorithm}
\usepackage{algpseudocode}
\usepackage{pbalance}

\algnewcommand\algorithmicforeach{\textbf{for each}}
\algdef{S}[FOR]{ForEach}[1]{\algorithmicforeach\ #1\ \algorithmicdo}

\title{FLASH-D: FlashAttention with Hidden \\Softmax Division}
\author{%
\IEEEauthorblockN{Kosmas Alexandridis, Vasileios Titopoulos, Giorgos Dimitrakopoulos}
\IEEEauthorblockA{Integrated Circuits Lab, 
Electrical and Computer Engineering, 
Democritus University of Thrace, Xanthi, Greece}}

\begin{document}

\maketitle

\begin{abstract}
The transformer's attention mechanism has revolutionized AI and machine learning, with its efficient computation being crucial to its performance. However, calculating attention involves matrix operations interspersed with softmax rescaling, which inherently slows down computation and requires processing the entire input sequence. Building on online softmax computation, FlashAttention integrates softmax calculation with matrix arithmetic, enabling tiled computation independent of sequence length. While optimized for GPUs, FlashAttention's simplicity makes it amenable to direct hardware acceleration. This work re-evaluates the core FlashAttention kernel, presenting \emph{FLASH-D} a \emph{mathematically equivalent}, yet simplified, formulation that achieves: (a) hiding softmax division within other non-linear function evaluations; (b) inherently numerically stable computation of exponentials, eliminating the need for maximum value subtraction; and (c) a reduction in computational cost without introducing numerical approximations to the FlashAttention kernel. Importantly, the essential FlashAttention properties that facilitate efficient tiled implementation are fully preserved. Hardware implementation results at 28nm demonstrate that this proposed formulation achieves a 22.8\% reduction in area and a 20.3\% reduction in power, on average, compared to state-of-the-art parallel hardware architectures without any performance penalty.
\end{abstract}


\section{Introduction}
Current state-of-the-art ML and AI systems are characterized by deep learning models with billions of parameters, achieving human-level performance in tasks like image recognition~\cite{vit} and natural language processing~\cite{deepseek}. A key innovation driving this progress is the attention mechanism~\cite{base_attn}, which allows models to focus on relevant parts of the input data.
This has led to breakthroughs like transformer networks and large language models that are revolutionizing how machines understand and interact with the world.

The increasing demand for processing long sequences in transformer models has exposed the complexity of the attention mechanism~\cite{longformer}. This computational burden, makes inference (and training) prohibitively expensive for extended context lengths. Traditional attention calculates pairwise similarities between all tokens, leading to a massive number of operations and memory traffic. 

To reduce the number of operations either by approximating the full attention matrix or by selectively focusing on the most relevant parts of the input sequence, techniques such as sparse attention~\cite{sparse_attn}, linear attention~\cite{lin_attn}, and low-rank attention~\cite{low_rank_attn} have been explored, each balancing accuracy and computational efficiency.

Early attention hardware accelerators optimize which attention-relevant matrices, such as key, value and queries, remain stationary in local SRAMs and which are streamed to compute the attention scores~\cite{a3,keller,lu}. If the accelerator cannot buffer and compute intermediate results for the entire sequence length, these intermediate results are written to back memory, which negatively impacts performance. To decouple the accelerator's computational resources and local memory from the sequence length, the designs of~\cite{lazy_softmax,cosa} optimize key components of attention such as matrix arithmetic and softmax function evaluation. 
In addition to data-flow and computation reordering optimizations, other techniques exploit token similarity~\cite{elsa,tsacc} to 
skip unnecessary computation and to reduce latency and power consumption. In-memory computation~\cite{xformer} has also been explored for optimizing computation of attention.

In parallel, FlashAttention~\cite{fa, fa2, nsquared}, originally developed for GPUs, has emerged as a powerful technique for accelerating attention. By leveraging tiling and fusing online softmax computation with matrix arithmetic, it enables parallel execution and reduces memory traffic. These features allow FlashAttention to lower execution latency and simplify the processing of long sequences without compromising accuracy. 

In this work, we aim at preserving the favorable IO properties of FlashAttention and \emph{simplify its core computation kernel}. To do this, 
we \emph{rewrite} the \emph{forward pass} of FlashAttention kernel used in transformer inference in a \emph{mathematical equivalent form}, called \emph{FLASH-D}, enabling several favorable computational features:
\begin{itemize}
\item
The softmax division is hidden within the reformulated non-linear function evaluations. This eliminates the need for explicit division, either during computation~\cite{fa} or as a post-processing step when following the lazy softmax approach~\cite{lazy_softmax,nsquared}.
\item
Without altering the algorithm or introducing any approximation,  softmax is replaced by an equivalent sigmoid function of attention score differences that is inherently numerically stable. This transformation improves parallelism and removes the requirement to scale inputs by the maximum attention score, which is necessary otherwise for numerically stable softmax computation~\cite{koca}.
\item
New possibilities emerge for reducing multiplications or memory accesses while preserving the core features of FlashAttention, which enable tiled computation to enhance locality and decrease memory traffic~\cite{fa,fa2}.
\end{itemize}

To assess the efficiency of FLASH-D for designing FlashAttention-based hardware accelerators, we implemented the optimized FlashAttention2 kernel and FLASH-D using a fully unrolled systolic architecture, in 28nm ASIC technology. Experimental results show that the proposed approach achieves significant area and power savings, ranging from 20--28\% and 16--27\%, respectively, for various hidden dimension sizes and floating-point number formats when compared to parallel hardware implementation of the kernel of FlashAttention2 state-of-the-art algorithm under the same performance.

\section{FlashAttention-based Hardware Accelerators}
Attention is a fundamental concept in machine learning, particularly in deep learning models that utilize the transformer architecture~\cite{gpt2,fastervit}.  It guides transformer models to take into account only relevant context: A user query is compared to a set of key vectors leading to an attention score matrix which is used to retrieve the information the user requested from a set of value vectors. 
In practice, attention is applied across multiple heads in parallel~\cite{base_attn}, allowing the model to comprehend more complex relationships. 
Without loss of generality, we limit our descriptions to single-head attention.

\subsection{Attention Kernel}
For a query $\vec{q}$ and 
a set of key and value vectors 
$K=\vec{k}_1,\ldots, \vec{k}_N$ and 
$V=\vec{v}_1, \ldots, \vec{v}_N$ attention is defined as 
\begin{equation}
s_i = \text{dot}(\vec{q}, \vec{k}_i)\qquad f_i = \frac{e^{s_i}}{\sum_j e^{s_j}}\qquad \text{Attn}(\vec{q}, \text{K}, \text{V}) = \sum_i f_i\, \vec{v}_i
\nonumber
\end{equation}

The attention score $s_i$ represents the similarity between a given query and the $i$th key vector computed via a dot product. To identify the most relevant tokens, the softmax function is applied to all scores corresponding to the same query vector. Softmax first exponentiates each score and then divides it by the sum of all exponentials. The final output is obtained by multiplying the attention scores of a query by all value vectors, where the contribution of each value to the output is determined by its corresponding attention score normalized via softmax.

Exponentiating scores can cause infinite values that would affect the final result. To prevent this, \emph{safe softmax} subtracts the maximum score from all scores, thus avoiding overflow while keeping its core properties
$f_i = e^{s_i-\max}/\sum_j e^{s_j-\max}$.

\subsection{Baseline FlashAttention Algorithm}
Algorithmically, attention calculation faces one major
bottleneck. The softmax operation should be applied across the entire sequence length. This requirement reduces the extent to which attention can be parallelized or limits the application of attention to shorter sequence lengths~\cite{consmax}.

To overcome this limitation, FlashAttention~\cite{fa}, driven by the online calculation of softmax function~\cite{online-softmax}, reorganized computations involved in attention kernel and enabled arbitrary tiling that reduced also memory traffic. The forward pass of baseline FlashAttentionm, which is the focus of this work, is shown in Alg.~\ref{alg:flash-attn} using vector-oriented operations. An equivalent block-based definition can be found in~\cite{fa}.

\begin{algorithm}[h]
\caption{FlashAttention}\label{alg:flash-attn}
\begin{algorithmic}[1]
\ForEach {query $\vec{q}$}
\For{$i = 1:N$} 
\State $s_i \gets \text{dot}(\vec{q}, \vec{k}_i)$
\State $m_i \gets \max(m_{i-1}, s_i)$
\State $\ell_i \gets \ell_{i-1}e^{m_{i-1}-m_i}+e^{s_i-m_i}$
\State $\vec{o}_i \gets
\vec{o}_{i-1} \left(\frac{\ell_{i-1}e^{m_{i-1}-m_i}}{\ell_i}\right)+
\vec{v}_i \left(\frac{e^{s_i-m_i}}{\ell_i}\right)$
\EndFor
\State $\text{Attn}(\vec{q}, \text{K}, \text{V}) \gets \vec{o}_N$ 
\EndFor
\end{algorithmic}
\end{algorithm}

At each iteration, the dot product of the query vector and a key vector yields a similarity score, denoted as $s_i$. Subsequently, $m_i$ holds the current maximum similarity score, while $\ell_i$ incrementally accumulates the sum of the exponentials of each $s_i$ minus the present maximum score. The multiplication by $e^{m_{i-1}-m_i}$ in the calculation of $\ell_i$
adjusts the prior maximum value used whenever the current maximum
$m_i$ differs from the previous maximum $m_{i-1}$. Similarly, the output vector $\vec{o}_i$ is updated by adding the new value vector $\vec{v}_i$ weighted by its softmax importance, to the adjusted preceding output vector, $\vec{o}_{i-1}$. The final attention vector for one query vector is returned in $\vec{o}_N$.

\subsection{Flash Attention with Lazy Softmax}
Instead of calculating the exponential sum and division before the multiplication with the value vectors, lazy softmax architectures~\cite{lazy_softmax, elsa} accumulate the
weighted sum and the exponent sum in parallel and perform
division as a final step. To avoid the expensive division
operation other approaches compute softmax in the logarithmic domain~\cite{cosa, edgebert}.
\begin{equation}
s_i\!=\! \text{dot}(\vec{q}, \vec{k}_i)\quad f_i\!=\! e^{s_i-\max}\quad 
\text{Attn}(\vec{q}, \text{K}, \text{V})\!=\! \frac{\sum_i f_i \vec{v}_i}{\sum_j e^{s_j-\max}}\nonumber
\end{equation}
This postponed division approach has been adopted by FlashAttention2 algorithm~\cite{fa2} that is shown in Alg.~\ref{alg:flash-attn2}. 
\begin{algorithm}
\caption{FlashAttention2: Lazy Softmax Division}\label{alg:flash-attn2}
\begin{algorithmic}[1]
\ForEach {query $\vec{q}$}
\For{$i = 1:N$} 
\State $s_i \gets \text{dot}(\vec{q}, \vec{k}_i)$
\State $m_i \gets \max(m_{i-1}, s_i)$
\State $\ell_i \gets \ell_{i-1}e^{m_{i-1}-m_i}+e^{s_i-m_i}$
\State $\vec{o}_i \gets \vec{o}_{i-1} e^{m_{i-1}-m_i}+\vec{v}_i e^{s_i-m_i}$
\EndFor
\State $\text{Attn}(\vec{q}, \text{K}, \text{V}) \gets \vec{o}_N/\ell_N$
\EndFor
\end{algorithmic}
\end{algorithm}

FlashAttention2 keeps the same basic principle of operation of baseline FlashAttention but removes all internal division operations (line 6 of Alg.~\ref{alg:flash-attn}) and replaces them with one final vector division at the end (line 8 of Alg.~\ref{alg:flash-attn2}). In this way, it leads to a simpler design that can be more easily parallelized and implemented directly in hardware. 

The FlashAttention2 kernel shown in Alg.~\ref{alg:flash-attn2} involves two for loops that can be unrolled to enhance parallelism and computational throughput. While unrolling the inner loop is a valid option, it still maintains the serial dependency across the three variables that constitute attention's internal state, namely $m_i, \ell_i$, and $\vec{o}_i$. A more scalable approach is to unroll the outer loop, allowing the FlashAttention2 kernel to process multiple query vectors in parallel within the same blocks of key and value vectors. In this case, the internal state is kept independently for each query vector, thus eliminating serial dependencies.

\begin{figure}
\centering
\includegraphics[width=0.92\columnwidth]{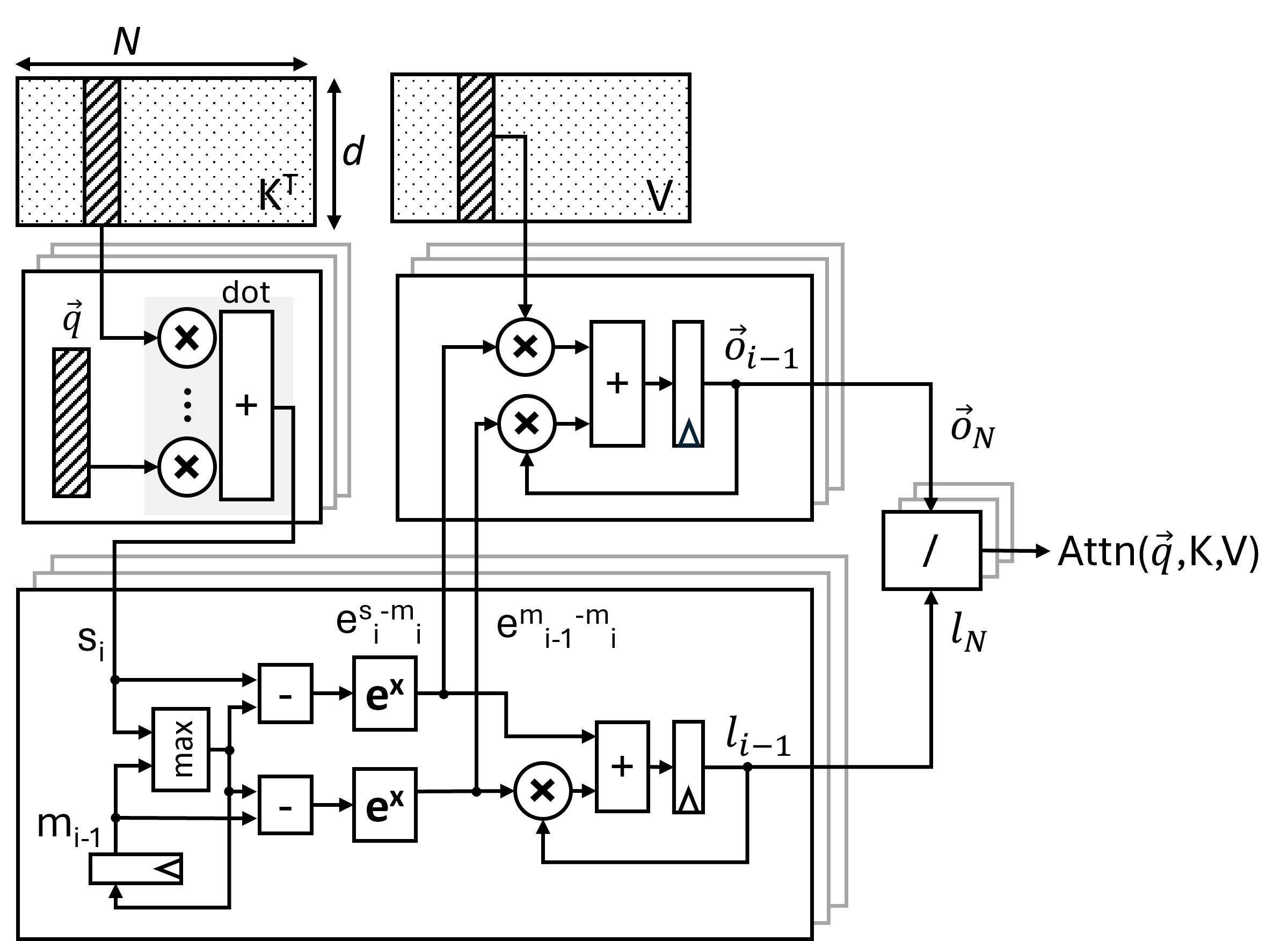}
\caption{A parallel hardware architecture for FlashAttention2 kernel for multiple preloaded query vectors.}
\label{f:flashattn2-hw}
\end{figure}

Fig.~\ref{f:flashattn2-hw} depicts the parallel hardware structure that corresponds to the outer-loop unrolling architecture of FlashAttention2. In this setup, a block of query vectors is loaded locally, while the key vectors are sequentially provided to each block for computing the dot products~\cite{ol-align,dion}. This results in the determination of a new maximum value and an updated running sum of exponents. The value vectors are streamed into the accelerator, allowing the output vectors for different query vectors to be updated simultaneously. After all key and value vectors are processed, a final division operation is performed to compute the attention for each query vector. The process concludes once all query vectors have been processed.

FlashAttention2 eliminates the need for a full softmax hardware unit, as the exponentiations and final division are computed independently. Several methods have been proposed for performing these two non-linear operations in hardware. These approaches include piece-wise linear approximations following range reduction~\cite{koca}, logarithmic quantization-based transformations~\cite{sole}, and other approximations~\cite{exp2two} that convert exponential functions into simpler powers of two, allowing for shift-and-add algorithms.

\section{FlashAttention with Hidden Softmax Division}
The two algorithmic variants of FlashAttention perform identical computations. Their sole difference lies in the timing of the softmax division. In baseline FlashAttention, the division is executed incrementally during output accumulation, whereas in FlashAttention2, it is rescheduled to the end of the computation. 

In this work, we aim at rewriting baseline FlashAttention to effectively \textbf{hide the softmax division} within non-linear function evaluations, without resorting to any approximations. By doing so, we intend to introduce a novel, yet equivalent, algorithmic variant of FlashAttention that facilitates energy-efficient hardware accelerators. We accomplish this transformation in two steps.

\subsection{Redefine output recursion as a weighted contribution of value vectors}
First, we need to rewrite the recursive output computation of baseline FlashAttention. From line 5 of Alg.~\ref{alg:flash-attn} we know that the output is computed as follows: 
\begin{equation}
\vec{o}_{i} = 
\vec{o}_{i-1} \left(\frac{\ell_{i-1}e^{m_{i-1}-m_i}}{\ell_i}\right)+
\vec{v}_i \left(\frac{e^{s_i-m_i}}{\ell_i}\right)\\
\label{e:flash-out}
\end{equation}

\noindent From the definition of $\ell_i = \ell_{i-1}e^{m_{i-1}-m_i}+e^{s_i-m_i}$ in line 4 of Alg.~\ref{alg:flash-attn} we get that:
\begin{equation*}
\ell_{i-1}e^{m_{i-1}-m_i} = \ell_i - e^{s_i-m_i}
\end{equation*}
Replacing this term to the output recursion~\eqref{e:flash-out} we get that
\begin{align}
\vec{o}_i & =
\vec{o}_{i-1} \left(\frac{\ell_i - e^{s_i-m_i} }{\ell_i}\right)+
\vec{v}_i \left(\frac{e^{s_i-m_i}}{\ell_i}\right)\nonumber\\
 & =
\vec{o}_{i-1} \left(1 - \frac{e^{s_i-m_i}}{\ell_i}\right)+
\vec{v}_i \left(\frac{e^{s_i-m_i}}{\ell_i}\right)
\label{e:inter}
\end{align}
Setting 
\begin{equation}
w_i = \frac{e^{s_i-m_i}}{\ell_i} 
\label{e:w-basic}
\end{equation}
and replacing it in~\eqref{e:inter} we get an equivalent, but simplified form, for the recursive computation of the output in baseline FlashAttention: 
\begin{equation}
\boxed{\vec{o}_i = \vec{o}_{i-1} (1 - w_i)+ \vec{v}_i w_i}
\label{e:out-final}
\end{equation}

This newly rewritten form of~\eqref{e:flash-out} reveals that, effectively, the output is a weighted sum of the previously accumulated output, $\vec{o}_{i-1}$, and the new value vector, $\vec{v}_i$. The contribution of each part is determined by the value of $w_i$ which, according to~\eqref{e:w-basic}, corresponds to an incrementally formed softmax since it includes the exponent of each attention score divided by the running sum of exponents accumulated in $\ell_i$. Thus,  
by definition, $w_i$ is always positive and less than 1.

\subsection{Recursive weight computation}
Next, we aim to establish a recursive relation that links the current weight, 
$w_i$ to the weight from the previous iteration, 
$w_{i-1}$ and the corresponding attention scores. Our goal is to use the planned recursive relation \emph{exclusively}, \emph{replacing} the recursive computation of $\ell_i$ and $m_i$, as well.

To do so, based on the definition of weight $w_i$ in~\eqref{e:w-basic}, we can equivalently write that $w_{i-1} = e^{s_{i-1}-m_{i-1}}/\ell_{i-1}$. Solving for $\ell_i$ and $\ell_{i-1}$, respectively, we get that:
\begin{equation}
\ell_{i} = \frac{e^{s_{i}-m_{i}}}{w_{i}}\qquad
\ell_{i-1} = \frac{e^{s_{i-1}-m_{i-1}}}{w_{i-1}}
\label{e:l-defs}
\end{equation}

Replacing $\ell_i$ and $\ell_{i-1}$ of~\eqref{e:l-defs} in the recursive equation used to compute the sum of exponents $\ell_i = \ell_{i-1}e^{m_{i-1}-m_i}+e^{s_i-m_i}$ (repeated from line 4 of Alg.~\ref{alg:flash-attn}) we get that:
\begin{equation}
\frac{e^{s_i-m_i}}{w_i} = 
\frac{e^{s_{i-1}-m_{i-1}}}{w_{i-1}}e^{m_{i-1}-m_i}+
e^{s_i-m_i}
\end{equation}
Simple algebraic manipulation allows us to remove the contribution of $m_{i-1}$ leading to:
\begin{equation}
\frac{e^{s_i-m_i}}{w_i} = 
\frac{e^{s_{i-1}-m_i}}{w_{i-1}}+
e^{s_i-m_i}
\label{e:dum}
\end{equation}
The term $e^{-m_i}$ is common to all terms in~\eqref{e:dum} and it can be simplified leading to:
\begin{equation}
\frac{e^{s_i}}{w_i} = \frac{e^{s_{i-1}}}{w_{i-1}}+
e^{s_i}
\label{e:first-w-rec}
\end{equation}
At this point, we have expressed $w_i$ recursively, depending only on the previous weight $w_{i-1}$ and neighbor attention scores
$s_i$ and $s_{i-1}$ of the same query vector with respect to two consecutive key vectors. Furthermore, the need to compute the running maximum attention score is also removed. In Section~\ref{ss:proposed}, we will show that this has no negative effects on the numerical stability of computing $w_i$.

To clarify the structure of $w_i$ in~\eqref{e:first-w-rec}, we first factor out $e^{s_i}$ and simplify both sides. This transforms~\eqref{e:first-w-rec} into
\begin{equation}
\frac{1}{w_i} = \frac{e^{s_{i-1}-s_i}}{w_{i-1}}+ 1
\nonumber
\end{equation}
rearranging the above fraction yields
\begin{equation}
w_i = \frac{w_{i-1}}{w_{i-1} + e^{s_{i-1}-s_i}}.
\label{e:w}
\end{equation}
According to~\eqref{e:w-basic}, $w_i$ is always positive. Therefore, we can replace $w_{i-1}$ in~\eqref{e:w} with $e^{\ln w_{i-1}}$ and rewrite it as follows:
\begin{align}
w_i & = \frac{e^{\ln w_{i-1}}}{e^{\ln w_{i-1}}+e^{s_{i-1}-s_i}}
    = \frac{1}{1 + e^{s_{i-1}-s_i-\ln w_{i-1}}} \nonumber\\
    & = \frac{1}{1 + e^{-(s_i-s_{i-1}+\ln w_{i-1})}} 
\label{e:w-b}
\end{align}
Eq.~\eqref{e:w-b} reveals that weight $w_i$ is the result of applying sigmoid function $\sigma(x) = 1/(1+e^{-x})$ to the difference of attention scores $s_i-s_{i-1}$ achieved by the same query vector for consecutive key vectors incremented (skewed) by the natural logarithm of previous weight $w_{i-1}$. In other words, 
\begin{equation}
\boxed{w_i = \sigma(s_i-s_{i-1}+\ln w_{i-1})}
\label{e:w-last}
\end{equation}

\subsection{FLASH-D: Hiding Softmax Division in FlashAttetion}
\label{ss:proposed}
The sigmoid-based computation of the newly defined weights in~\eqref{e:w-last} and the revised recursive computation of the output derived in~\eqref{e:out-final} facilitates a more compact reformulation of the FlashAttention kernel that is shown in Alg.~\ref{alg:proposed}.
For each attention score $s_i$, a new weight $w_{i}$ is computed based on previous attention score $s_{i-1}$ and weight $w_{i-1}$, that in turn determines how much the new value vector $\vec{v}_i$ would affect the accumulation at the output $\vec{o}_i$. In the first iteration, the weight is set  equal to 1.

The output is gradually formed following the FlashAttention paradigm. For instance, following the evolution of the inner loop of Alg.~\ref{alg:proposed}, $w_1$ is set to 1 that leads to $\vec{o}_1 = \vec{v}_1$. The first value vector $\vec{v}_1$ will be weighted by its corresponding exponential in the next iteration. Specifically, in the next iteration, $w_2 = 1/(1+e^{-(s_2-s_1)})=e^{s_2}/(e^{s_1}+e^{s_2})$, which gives $1-w_2 = e^{s_1}/(e^{s_1}+e^{s_2})$. These coefficients lead to an incremental output $\vec{o}_2 = e^{s_1}\vec{v}_1/(e^{s_1}+e^{s_2})+
e^{s_2}\vec{v}_2/(e^{s_1}+e^{s_2})$. As the iterations progress each $\vec{v}_i$ is multiplied to the appropriate exponential, while the sum of exponentials is growing including more terms $e^{s_i}$ at the denominator of each fraction. Computation is completed in $N$th step with $\vec{o}_N$, when all key and value vectors have been processed.

\begin{algorithm}[t]
\caption{FLASH-D: Division hidden in sigmoid function}\label{alg:proposed}
\begin{algorithmic}[1]
\ForEach {query $\vec{q}$}
\For{$i = 1:N$} 
\State $s_i \gets \text{dot}(\vec{q}, \vec{k}_i)$
\If{$i\ne 1$}
    \State $w_i \gets \sigma(s_i-s_{i-1}-\ln w_{i-1})$
\Else
    \State $w_i \gets 1$
\EndIf
\State $\vec{o}_i \gets \vec{o}_{i-1} (1 - w_i)+\vec{v}_i\, w_i$
\EndFor
\State $\text{Attn}(\vec{q}, \text{K}, \text{V}) \gets \vec{o}_N$
\EndFor
\end{algorithmic}
\end{algorithm}

The proposed Alg.\ref{alg:proposed} is \textbf{a one-to-one equivalent of the baseline FlashAttention} (Alg.\ref{alg:flash-attn}), derived through mathematical reformulation without introducing any approximations at any stage. The incremental \textbf{division by the sum of exponents} in the baseline FlashAttention is effectively \textbf{hidden within the sigmoid function}, thereby \textbf{merging it with the corresponding exponential function evaluations}. 

Even without reducing each attention score $s_i$ by the maximum attention score across all key vectors, \textbf{the numerical stability of FLASH-D is guaranteed}. 
This is because we can completely avoid situations that would lead to exponential overflow and instead return the default correct values for $w_i$. In fact, the cases that could cause exponential overflows correspond to a range of attention score differences $s_i-s_{i-1}$, where computing the weight function becomes meaningless.

\begin{figure}[thb]
\centering
\includegraphics[width=0.98\columnwidth]{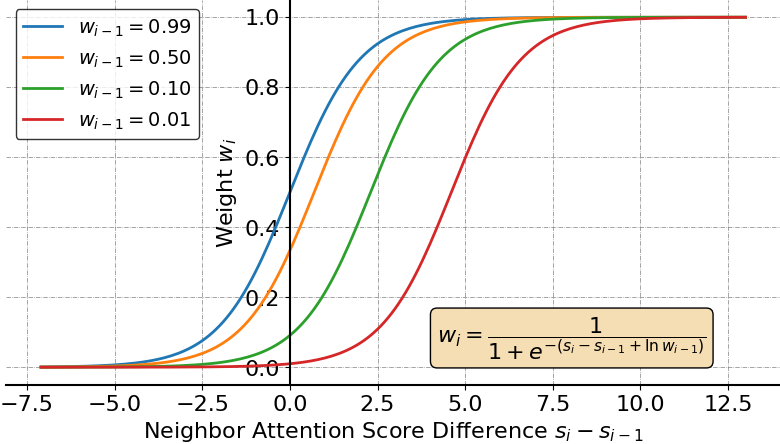}
\caption{Weight $w_i$ function for various values of consecutive attention score differences $s_i-s_{i-1}$. The four weight graphs correspond to four different values of the previous weight $w_{i-1}$.}
\label{f:sigmoid-w}
\end{figure}

To clarify this argument we plot in Fig.~\ref{f:sigmoid-w} weight $w_i$ for different attention score differences $s_i-s_{i-1}$. The graphs correspond to four distinct values of the previous weight. 
Since $w_i$ is derived using the sigmoid function, its dynamic range is always between 0 and 1. The leftmost graph represents $w_{i-1}=0.99$, which closely follows the standard sigmoid function, as $\ln w_{i-1}$ is effectively zero. As $w_{i-1}$  decreases, the weight function shifts to the right. In all cases, it is evident that when $s_i-s_{i-1}$ falls outside the range [-6,11], there is no need to compute the weight function explicitly. In such cases, $w_i$ will be very close to either 0 or 1, allowing it to be set directly to the smallest or largest values within $(0,1)$ by default. Consequently, for values of $s_i-s_{i-1}$ outside the range [-6,11], the exponential within the sigmoid function is skipped, thus preventing any overflow condition. 

Additionally, in such cases, \textbf{output update can be simplified}. When $s_i-s_{i-1} \le -6$, $w_i$ approaches 0, meaning the output vector $\vec{o}_i$ remains unchanged (see line 9 of Alg.~\ref{alg:proposed}), eliminating the need 
of loading of $\vec{v}_i$ and performing any
for vector multiplication or addition. Conversely, when $s_i-s_{i-1} \ge 11$, $w_i$ approaches 1 by default, causing the output vector to effectively ``forget'' previous value contributions and update only with the new value vector $\vec{v}_i$. In this scenario, multiplication and addition operations are also bypassed.

\section{Hardware Architecture of FLASH-D}
Flash-D preserves all the characteristics of FlashAttention variants while embedding the softmax division within the sigmoid non-linear function, ensuring numerical stability is maintained. This transformation greatly simplifies the hardware implementation of FLASH-D.

\subsection{Overall Organization}
\begin{figure}[t]
\centering
\includegraphics[width=0.9\columnwidth]{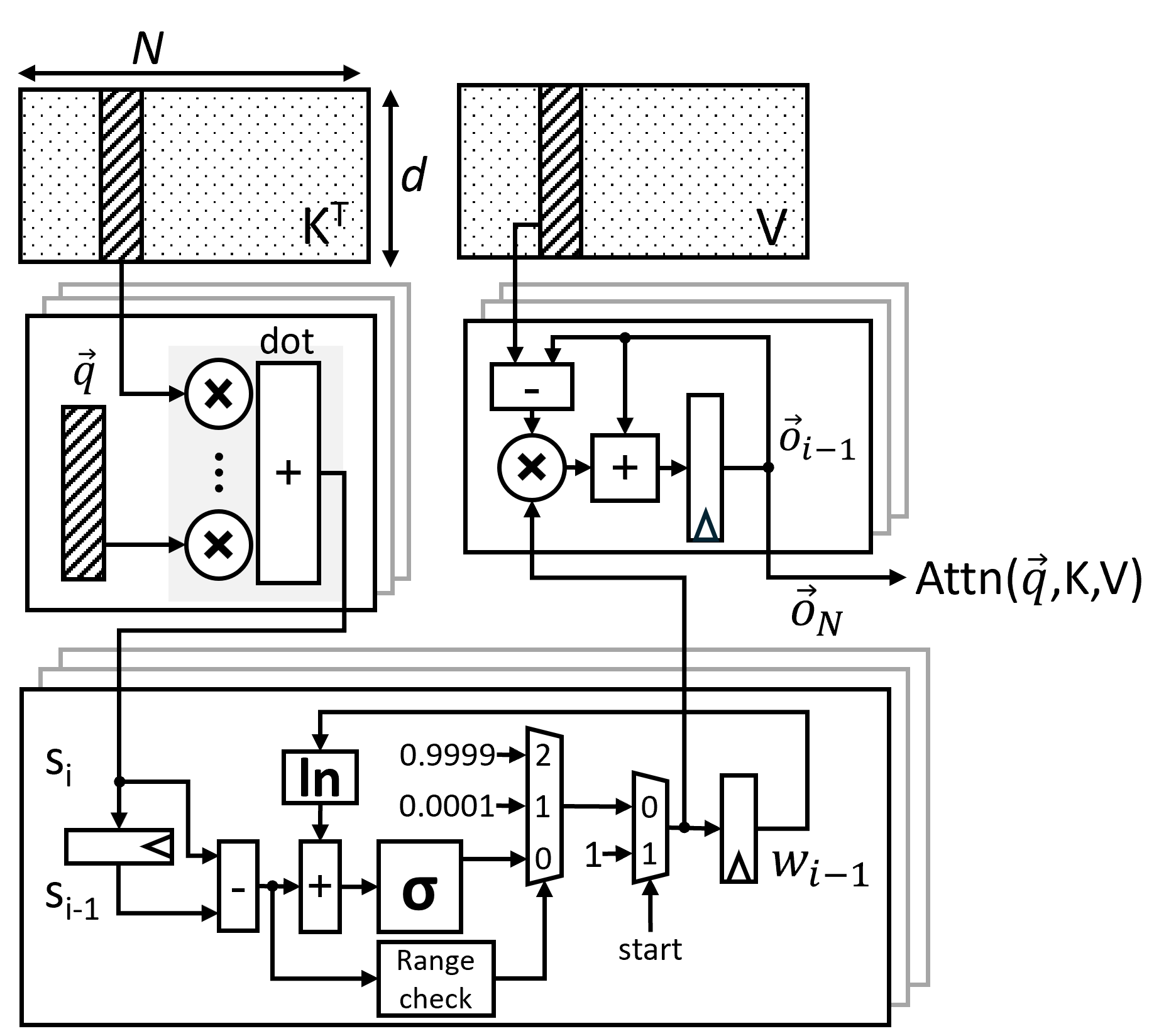}
\caption{A parallel hardware architecture for FLASH-D kernel for multiple preloaded query vectors.}
\label{f:flashd-hw}
\end{figure}

As illustrated in Fig.\ref{f:flashd-hw}, the hardware design of FLASH-D follows the same overall architecture of FlashAttention2 (Fig.\ref{f:flashattn2-hw}). Both approaches share the same dataflow, processing multiple query vectors in parallel in an unrolled manner, driven by the same key and value vectors. The primary differences lie in three key areas.

First, the division at the output of FlashAttention2 kernel is not required in Flash-D. While division is neither removed nor approximated, it is implemented incrementally during the computation of each weight $w_i$ for each query vector.

Second, the running sum-of-exponents $\ell_i$ and the maximum attention score $m_i$ used in FlashAttention2, as shown in Fig.~\ref{f:flashattn2-hw}, are eliminated. The running sum-of-exponents $\ell_i$ is implicitly embedded in the computation of each weight (see Eq.~\eqref{e:w-basic}), so it does not need to be explicitly computed. Additionally, the maximum value is no longer necessary, as numerical stability is maintained by ensuring that the attention score difference remains within the active region of [-6,11], where the sigmoid function can be effectively computed.

Third, the output computation module in the FLASH-D hardware requires one vector adder, one subtractor, and one multiplier, compared to the two multipliers and one adder found in FlashAttention2. This hardware simplification, where one multiplier is replaced by a subtractor, is made possible by the new weighted definition of the output recursion (line 9 of Alg. 3), which can be equivalently written as follows:
\begin{equation}
\vec{o}_i =\vec{o}_{i-1} (1 - w_i) + \vec{v}_i\, w_i =  \vec{o}_{i-1} + (\vec{v}_i - \vec{o}_{i-1})\, w_i
\end{equation}

\subsection{Non-linear function evaluation in FLASH-D}
The non-linear function evaluations in FLASH-D involve the sigmoid and natural logarithm functions, both of which are well-researched and have various hardware implementations available. Without focusing on any specific simplifications, this work implements both functions using standard piece-wise linear (PWL) approximations.

This design choice is well justified for both functions. The sigmoid function has a well-defined structure and output range, making it a natural fit for the PWL approach. Furthermore, in FLASH-D the input dynamic range is constrained to [-6, 11], so no computation is required outside this range.
The natural logarithm function is used solely to compute the natural logarithm of the previous weight, i.e., $\ln w_{i-1}$, which has a narrow input dynamic range of (0,1). As a result, there is no need to compute a generic logarithm; instead, we require one that consistently returns a negative result that follows the value of the previous weight. Again, the PWL approach proves to be a suitable choice for this case.

In both cases, we approximated the functions using 8 line segments. The coefficients of each segment are produced via  
{\tt pwlf} a Python-based PWL-fit optimization library~\cite{pwlf}. 

\section{Evaluation}
The experimental evaluation has two parts. First, we highlight the area and power benefits provided by the hardware implementation of FLASH-D (Fig.~\ref{f:flashd-hw}) compared to the optimized hardware implementation of the FlashAttention2 kernel shown in Fig.~\ref{f:flashattn2-hw}. Second, our goal is to quantify how often the weight and output update can be skipped in real LLM applications using FLASH-D as the forward pass without affecting the outcome of FlashAttention.

\subsection{Hardware complexity}
To evaluate the hardware complexity of the two designs, we implemented the main block shown in the foreground of Figs.~\ref{f:flashattn2-hw} and~\ref{f:flashd-hw} for various hidden dimension sizes ($d$) and for two reduced-precision floating-point formats: BFloat16~\cite{bfloat16} and FP8-E4M3~\cite{fp8}. In practice, the total cost of the unrolled hardware serving multiple query vectors in parallel will be the cost for one query multiplied by the number of parallel query vectors served. The query vectors are preloaded separately in the architecture, while the key and value vectors are loaded and broadcasted to all parallel blocks. For both FLASH-D and FlashAttention2 kernels, we assume that we can read from local memories in each cycle one key and one value vector of $d$ elements each.

Both hardware blocks were implemented in C++\footnote{Publicly available on https://github.com/ic-lab-duth/FLASH-D.git}
and synthesized into Verilog using Catapult HLS with a 28-nm standard-cell library. For verifying the correctness of the C++ code, we integrated it to {\tt llama2.c}~\cite{llama-repo} and we received exactly the same replies as the original implementation for all examined queries.
Both designs operate at the same pipelined latency with a clock frequency of 500 MHz. Latency depends on the size of the hidden dimension, requiring 8, 10, and 12 cycles for $d=\{16,64,256\}$ elements, respectively. Verilog was synthesized using the Cadence digital implementation flow, while power consumption was estimated with the PowerPro power analysis and optimization tool. The reported power consumption represents the average power measured after executing attention kernels for various Large Language Models and benchmarks from PromptBench.

\begin{figure}[t]
\centering
\includegraphics[width=\columnwidth]{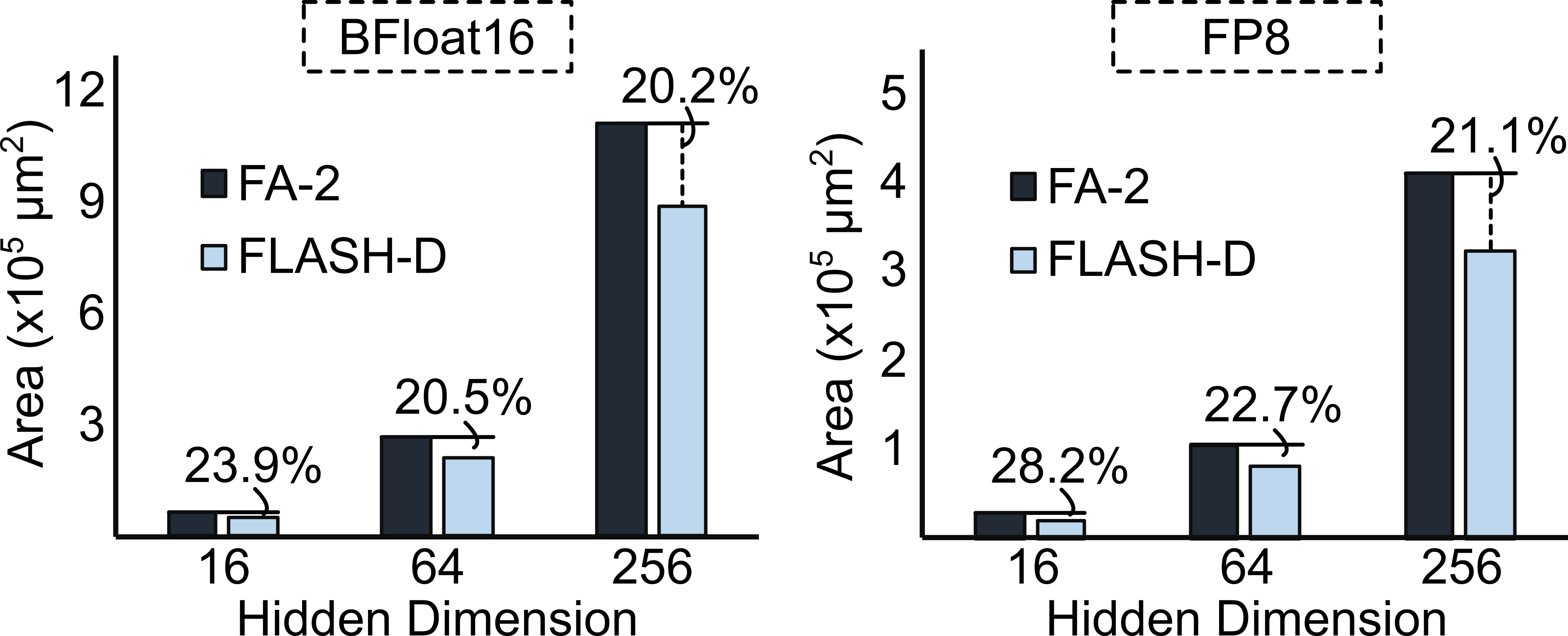}
\caption{The hardware area at 28 nm for FLASH-D and FlashAttention2 kernel for computing attention of a single query using BFloat16 and FP8-E4M3 floating-point formats, across different hidden dimension lengths.}
\label{f:area}
\end{figure}

\begin{figure}[t]
\centering
\includegraphics[width=\columnwidth]{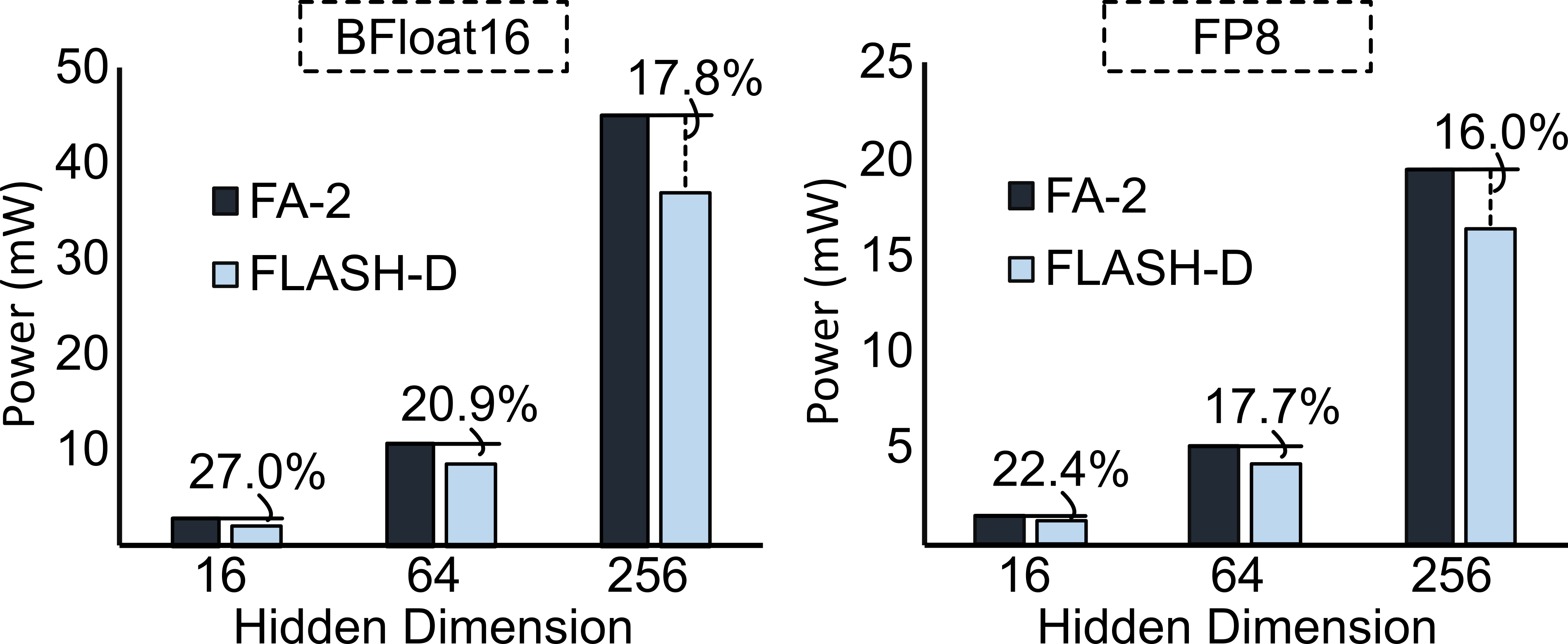}
\caption{The average power  for FLASH-D and FlashAttention2 kernel for computing attention of a single query using BFloat16 and FP8-E4M3 floating-point formats, across different hidden dimension lengths. Memory and IO power is not included since it is identical to both designs.}
\label{f:power}
\end{figure}

Figs.~\ref{f:area} and~\ref{f:power} show the area and power of the proposed FLASH-D hardware and the parallel hardware implementation of the FlashAttention2 computation kernel, for the two examined reduced precision floating-point formats and different sizes of the hidden dimension. Power estimation excludes memory power and focuses solely on the average power consumption of the computation kernel. The memory power in both cases is expected to be identical, as both approaches implement the same FlashAttention algorithm using the same computation order and data flows. The difference lies solely on how the computation kernel is executed internally.

As shown in Fig.~\ref{f:area}, FLASH-D reduces the hardware area by more than 20\% in all examined cases. These savings are a direct result of the restructured FlashAttention kernel. Standalone division is eliminated and fused within the sigmoid PWL function evaluation, effectively merging the exponential and division operations. One vector multiplier is saved in the output update module, and the sum-of-exponents and maximum logic are entirely removed.

The reduction in hardware complexity also improves power consumption, which is reduced by more than 16\% on average across all examined cases. Power savings are expected to increase further, as the computation-skipping criterion developed in FLASH-D would save additional memory power, which has not been quantified in the presented analysis.

\subsection{How often output update can be simplified}
To quantify how often the attention score differences fall outside the range 
[-6,11] described in Section~\ref{ss:proposed}, and thereby simplify the output update in FLASH-D (line 9 in Alg.\ref{alg:proposed}), we implemented the FLASH-D kernel in Python and integrated it into the forward pass of various contemporary LLM models available on HuggingFace~\cite{hf}. We then performed inference on these LLMs using the utilities and benchmarks provided by Microsoft's PromptBench workflow~\cite{promptbench}. The results for each case are summarized in Table~\ref{t:llm_bench}.

In all cases, there is small percentage of cases that output update can be simplified with either keeping the previous computed output or loading the new value vector without any further calculations. This percentage, even if small, is always a win scenario and does not represent any tradeoff across energy savings vs application-level performance. 
In the future, we plan to replace this pessimistic and static range check with an adaptive criterion that includes both the range of attention score differences and the value of the previous weight to decide when output computation can be simplified.

\begin{table}[t]
\caption{Percentage of skipped output updates during inference on different NLP benchmarks}
\label{t:llm_bench}
\begin{adjustbox}{width=\columnwidth}
\begin{tabular}{|c||c|c|c|c|c|c|}
  \hline
      \multirow{3}{*}{\textbf{LLM}} & \multicolumn{6}{c|}{\textbf{Benchmarks}} \\\cline{2-7}
      ~ & \multirow{2}{*}{\textbf{CSQA}} & \multirow{2}{*}{\textbf{GSM8K}} & \multirow{2}{*}{\textbf{QASC}} & \multirow{2}{*}{\textbf{MMLU}} & \multirow{2}{*}{\textbf{Date}} & \textbf{Object} \\ ~ & ~ & ~ & ~ & ~ & ~ & \textbf{Tracking}  \\ \hline
      Microsoft / & \multirow{2}{*}{0.8\%} & \multirow{2}{*}{1.7\%} & \multirow{2}{*}{2.2\%} & \multirow{2}{*}{2\%} & \multirow{2}{*}{1.5\%} & \multirow{2}{*}{2\%} \\
      Phi-3-mini-4k-instruct & ~ & ~ & ~ & ~ & ~ & ~ \\ \hline
      DeepSeek / & \multirow{2}{*}{2.5\%} & \multirow{2}{*}{2.0\%} & \multirow{2}{*}{2.2\%} & \multirow{2}{*}{2.7\%} & \multirow{2}{*}{2.4\%} & \multirow{2}{*}{2.8\%} \\ 
      Qwen-1.5B & ~ & ~ & ~ & ~ & ~ & ~ \\ \hline
      Meta / & \multirow{2}{*}{1.8\%} & \multirow{2}{*}{1.6\%} & \multirow{2}{*}{2.6\%} & \multirow{2}{*}{2.3\%} & \multirow{2}{*}{1.6\%} & \multirow{2}{*}{2.3\%} \\
      Llama-3.1-1B & ~ & ~ & ~ & ~ & ~ & ~ \\ \hline
      Google / & \multirow{2}{*}{1.2\%} & \multirow{2}{*}{0.5\%} & \multirow{2}{*}{0.51\%} & \multirow{2}{*}{1.4\%} & \multirow{2}{*}{0.8\%} & \multirow{2}{*}{0.83\%} \\
      Gemma2-2B & ~ & ~ & ~ & ~ & ~ & ~ \\ \hline
\end{tabular}
\end{adjustbox}
\end{table}

\section{Conclusions}
In this work, our target was to simplify the computational kernel of FlashAttention by hiding softmax division inside sigmoid non-linear function evaluation and reduce unnecessary multiplications in output computation. This reformulation achieved these goals without compromising numerical stability and without negatively affecting the favorable IO and memory-access properties of FlashAttention algorithm. The proposed approach reduces the area and power of the hardware computational kernel by 22.8\% and 20.3\%, on average, respectively, when compared to the parallel hardware implementation of FlashAttention2 state-of-the-art algorithm. Any other approach that simplifies attention mechanism is orthogonal to FLASH-D and can be applied for increasing efficiency further.

\bibliographystyle{IEEEtran}
\bibliography{refs}

\begin{thebibliography}{10}
\providecommand{\url}[1]{#1}
\csname url@samestyle\endcsname
\providecommand{\newblock}{\relax}
\providecommand{\bibinfo}[2]{#2}
\providecommand{\BIBentrySTDinterwordspacing}{\spaceskip=0pt\relax}
\providecommand{\BIBentryALTinterwordstretchfactor}{4}
\providecommand{\BIBentryALTinterwordspacing}{\spaceskip=\fontdimen2\font plus
\BIBentryALTinterwordstretchfactor\fontdimen3\font minus \fontdimen4\font\relax}
\providecommand{\BIBforeignlanguage}[2]{{%
\expandafter\ifx\csname l@#1\endcsname\relax
\typeout{** WARNING: IEEEtran.bst: No hyphenation pattern has been}%
\typeout{** loaded for the language `#1'. Using the pattern for}%
\typeout{** the default language instead.}%
\else
\language=\csname l@#1\endcsname
\fi
#2}}
\providecommand{\BIBdecl}{\relax}
\BIBdecl

\bibitem{vit}
A.~Dosovitskiy, L.~Beyer, A.~Kolesnikov, D.~Weissenborn, X.~Zhai, T.~Unterthiner, M.~Dehghani, M.~Minderer, G.~Heigold, S.~Gelly \emph{et~al.}, ``An image is worth 16x16 words: Transformers for image recognition at scale,'' \emph{arXiv preprint arXiv:2010.11929}, 2020.

\bibitem{deepseek}
D.~Guo, D.~Yang, H.~Zhang, J.~Song, R.~Zhang, R.~Xu, Q.~Zhu, S.~Ma, P.~Wang, X.~Bi \emph{et~al.}, ``Deepseek-r1: Incentivizing reasoning capability in llms via reinforcement learning,'' \emph{arXiv preprint arXiv:2501.12948}, 2025.

\bibitem{base_attn}
A.~Vaswani, N.~Shazeer, N.~Parmar, J.~Uszkoreit, L.~Jones, A.~N. Gomez, L.~Kaiser, and I.~Polosukhin, ``Attention is all you need,'' in \emph{Intern. Conf. on Neural Information Processing Systems (NIPS)}, 2017, p. 6000–6010.

\bibitem{longformer}
I.~Beltagy, M.~E. Peters, and A.~Cohan, ``Longformer: The long-document transformer,'' \emph{arXiv preprint arXiv:2004.05150}, 2020.

\bibitem{sparse_attn}
R.~Child, S.~Gray, A.~Radford, and I.~Sutskever, ``Generating long sequences with sparse transformers,'' \emph{arXiv preprint arXiv:1904.10509}, 2019.

\bibitem{lin_attn}
A.~Katharopoulos, A.~Vyas, N.~Pappas, and F.~Fleuret, ``Transformers are rnns: Fast autoregressive transformers with linear attention,'' in \emph{Intern. conference on machine learning}, 2020, pp. 5156--5165.

\bibitem{low_rank_attn}
N.~Tang, M.~Fu, K.~Zhu, and J.~Wu, ``Low-rank attention side-tuning for parameter-efficient fine-tuning,'' \emph{arXiv preprint arXiv:2402.04009}, 2024.

\bibitem{a3}
T.~J. Ham, S.~J. Jung, S.~Kim, Y.~H. Oh, Y.~Park, Y.~Song, J.-H. Park, S.~Lee, K.~Park, J.~W. Lee, and D.-K. Jeong, ``A3: Accelerating attention mechanisms in neural networks with approximation,'' in \emph{IEEE Intern. Symp. on High-Performance Computer Architecture (HPCA)}, 2020, p. 328–341.

\bibitem{keller}
B.~Keller, R.~Venkatesan, S.~Dai, S.~G. Tell, B.~Zimmer, C.~Sakr, W.~J. Dally, C.~T. Gray, and B.~Khailany, ``\BIBforeignlanguage{en}{A 95.6-{TOPS/W} deep learning inference accelerator with per-vector scaled 4-bit quantization in 5 nm},'' \emph{\BIBforeignlanguage{en}{IEEE Journal of Solid-State Circuits}}, vol.~58, no.~4, p. 1129–1141, 2023.

\bibitem{lu}
S.~Lu, M.~Wang, S.~Liang, J.~Lin, and Z.~Wang, ``Hardware accelerator for multi-head attention and position-wise feed-forward in the transformer,'' in \emph{IEEE Intern. System-on-Chip Conference (SOCC)}, 2020, pp. 84--89.

\bibitem{lazy_softmax}
H.~Jang, J.~Kim, J.-E. Jo, J.~Lee, and J.~Kim, ``Mnnfast: a fast and scalable system architecture for memory-augmented neural networks,'' in \emph{Intern. Symp. on Computer Architecture (ISCA)}, 2019, p. 250–263.

\bibitem{cosa}
Z.~Wang, G.~Wang, and G.~He, ``{COSA} plus: Enhanced co-operative systolic arrays for attention mechanism in transformers,'' \emph{IEEE Trans. on Computer-Aided Design of Integrated Circuits and Systems (TCAD)}, vol.~44, no.~2, p. 723–736, 2025.

\bibitem{elsa}
T.~J. Ham, Y.~Lee, S.~H. Seo, S.~Kim, H.~Choi, S.~J. Jung, and J.~W. Lee, ``{ELSA}: Hardware-software co-design for efficient, lightweight self-attention mechanism in neural networks,'' in \emph{Intern. Symp. on Computer Architecture (ISCA)}, 2021, p. 692–705.

\bibitem{tsacc}
Z.~Song, C.~Qi, Y.~Yao, P.~Zhou, Y.~Zi, N.~Wang, and X.~Liang, ``{TSAcc}: An efficient tempo-spatial similarity aware accelerator for attention acceleration,'' in \emph{{ACM/IEEE} Design Automation Conference}, 2024.

\bibitem{xformer}
S.~Sridharan, J.~R. Stevens, K.~Roy, and A.~Raghunathan, ``X-former: In-memory acceleration of transformers,'' \emph{IEEE Transactions on Very Large Scale Integration (VLSI) Systems}, vol.~31, no.~8, pp. 1223--1233, 2023.

\bibitem{fa}
T.~Dao, D.~Fu, S.~Ermon, A.~Rudra, and C.~R{\'e}, ``Flashattention: Fast and memory-efficient exact attention with {IO}-awareness,'' \emph{Advances in neural information processing systems}, vol.~35, pp. 16\,344--16\,359, 2022.

\bibitem{fa2}
T.~Dao, ``{Flashattention-2}: Faster attention with better parallelism and work partitioning,'' \emph{arXiv preprint arXiv:2307.08691}, 2023.

\bibitem{nsquared}
M.~N. Rabe and C.~Staats, ``Self-attention does not need {$O(n^2)$} memory,'' \emph{arXiv preprint arXiv:2112.05682}, 2021.

\bibitem{koca}
N.~A. Koca, A.~T. Do, and C.-H. Chang, ``Hardware-efficient softmax approximation for self-attention networks,'' in \emph{Intern. Symp. on Circuits and Systems (ISCAS)}, 2023, p. 1–5.

\bibitem{gpt2}
A.~Radford, J.~Wu, R.~Child, D.~Luan, D.~Amodei, I.~Sutskever \emph{et~al.}, ``Language models are unsupervised multitask learners,'' \emph{OpenAI blog}, p.~9, 2019.

\bibitem{fastervit}
A.~Hatamizadeh, G.~Heinrich, H.~Yin, A.~Tao, J.~M. Alvarez, J.~Kautz, and P.~Molchanov, ``Fastervit: Fast vision transformers with hierarchical attention,'' in \emph{Intern. Conf. on Learning Representations (ICLR)}, 2024.

\bibitem{consmax}
S.~Liu, G.~Tao, Y.~Zou, D.~Chow, Z.~Fan, K.~Lei, B.~Pan, D.~Sylvester, G.~Kielian, and M.~Saligane, ``Consmax: Hardware-friendly alternative softmax with learnable parameters,'' \emph{arXiv preprint arXiv:2402.10930}, 2024.

\bibitem{online-softmax}
M.~Milakov and N.~Gimelshein, ``Online normalizer calculation for softmax,'' \emph{arXiv preprint arXiv:1805.02867}, 2018.

\bibitem{edgebert}
T.~Tambe, C.~Hooper, L.~Pentecost, T.~Jia, E.-Y. Yang, M.~Donato, V.~Sanh, P.~Whatmough, A.~M. Rush, D.~Brooks, and G.-Y. Wei, ``{EdgeBERT:} sentence-level energy optimizations for latency-aware multi-task {NLP} inference,'' in \emph{Intern. Symposium on Microarchitecture (MICRO)}, 2021, p. 830–844.

\bibitem{ol-align}
K.~Alexandridis and G.~Dimitrakopoulos, ``Online alignment and addition in multiterm floating-point adders,'' \emph{IEEE Transactions on Very Large Scale Integration (VLSI) Systems}, vol.~33, no.~4, pp. 1182--1186, 2025.

\bibitem{dion}
D.~Filippas, C.~Nicopoulos, and G.~Dimitrakopoulos, ``Templatized fused vector floating-point dot product for high-level synthesis,'' \emph{Journal of Low Power Electronics and Applications}, vol.~12, no.~4, p.~56, 2022.

\bibitem{sole}
W.~Wang, S.~Zhou, W.~Sun, P.~Sun, and Y.~Liu, ``{SOLE:} hardware-software co-design of softmax and layernorm for efficient transformer inference,'' in \emph{{IEEE/ACM} Intern. Conference on Computer Aided Design (ICCAD)}, 2023, p. 1–9.

\bibitem{exp2two}
G.~C. Cardarilli, L.~Di~Nunzio, R.~Fazzolari, D.~Giardino, A.~Nannarelli, M.~Re, and S.~Spanò, ``A pseudo-softmax function for hardware-based high speed image classification,'' \emph{Scientific Reports}, vol.~11, 2021.

\bibitem{pwlf}
C.~F. Jekel and G.~Venter, \emph{{pwlf:} A Python Library for Fitting 1D Continuous Piecewise Linear Functions}, 2019.

\bibitem{bfloat16}
D.~Kalamkar, D.~Mudigere, N.~Mellempudi, D.~Das, K.~Banerjee, S.~Avancha, D.~T. Vooturi, N.~Jammalamadaka, J.~Huang, H.~Yuen \emph{et~al.}, ``A study of bfloat16 for deep learning training,'' \emph{arXiv preprint arXiv:1905.12322}, 2019.

\bibitem{fp8}
P.~Micikevicius, D.~Stosic, N.~Burgess, M.~Cornea, P.~Dubey, R.~Grisenthwaite, S.~Ha, A.~Heinecke, P.~Judd, J.~Kamalu \emph{et~al.}, ``Fp8 formats for deep learning,'' \emph{arXiv preprint arXiv:2209.05433}, 2022.

\bibitem{llama-repo}
A.~Karpathy, ``llama2.c,'' \url{https://github.com/karpathy/llama2.c.git}, 2023.

\bibitem{hf}
T.~Wolf, L.~Debut, V.~Sanh, J.~Chaumond, C.~Delangue, A.~Moi, P.~Cistac, T.~Rault, R.~Louf, M.~Funtowicz, J.~Davison, S.~Shleifer, P.~von Platen, C.~Ma, Y.~Jernite, J.~Plu, C.~Xu, T.~L. Scao, S.~Gugger, M.~Drame, Q.~Lhoest, and A.~M. Rush, ``Transformers: State-of-the-art natural language processing,'' in \emph{Conference on Empirical Methods in Natural Language Processing: System Demonstrations}, 2020, pp. 38--45.

\bibitem{promptbench}
K.~Zhu, Q.~Zhao, H.~Chen, J.~Wang, and X.~Xie, ``Promptbench: A unified library for evaluation of large language models,'' \emph{arXiv preprint arXiv:2312.07910}, 2023.

\end{thebibliography}
\end{document}